\pdfoutput=1

\documentclass[11pt]{article}

\usepackage[]{acl}

\usepackage{times}
\usepackage{latexsym}

\usepackage[T1]{fontenc}

\usepackage[utf8]{inputenc}

\usepackage{microtype}

\usepackage{inconsolata}

\usepackage{times}
\usepackage{latexsym}
\usepackage{amsmath}
\usepackage{amssymb}
\usepackage{graphicx}
\usepackage{subcaption}
\usepackage{multirow}
\usepackage{changepage}
\usepackage{hhline}
\usepackage{color}
\usepackage{booktabs}
\usepackage{bbm}
\usepackage{wrapfig,lipsum,booktabs}
\usepackage{makecell}
\usepackage{float}
\usepackage{arydshln}
\usepackage{enumitem}
\usepackage{dsfont}
\usepackage{bbding}
\usepackage{authblk}

\title{Toward Informal Language Processing: \\Knowledge of Slang in Large Language Models}

\renewcommand*{\Affilfont}{\normalsize\normalfont}
\makeatletter
\renewcommand\AB@affilsepx{ \protect\Affilfont}
\makeatother

\author[1]{Zhewei Sun}
\author[2]{Qian Hu}
\author[2]{Rahul Gupta}
\author[2,3]{Richard Zemel}
\author[1]{Yang Xu}
\affil[1]{University of Toronto, }
\affil[2]{Amazon Alexa AI, }
\affil[3]{Columbia University\protect\\} 

\affil[ ]{\ttfamily \{zheweisun, yangxu\}@cs.toronto.edu, \{huqia, gupra\}@amazon.com \protect\\}
\affil[ ]{\ttfamily zemel@cs.columbia.edu}

\begin{document}
\maketitle
\begin{abstract}


Recent advancement in large language models (LLMs) has offered a strong potential for natural language systems to process informal language. A representative form of informal language is slang, used commonly in daily conversations and online social media. To date, slang has not been comprehensively evaluated in LLMs   due partly to the absence of a carefully designed and publicly accessible benchmark.
Using movie subtitles, we construct a  dataset that supports evaluation on a diverse set of tasks pertaining to automatic processing of slang. For both evaluation and finetuning, we show the effectiveness of our dataset on two core applications: 1) slang detection, and 2) identification of regional and historical sources of slang from natural sentences. We also show how our dataset can be used to probe the output distributions of LLMs for interpretive insights.
We find that while LLMs such as GPT-4 achieve good performance in a zero-shot setting, smaller BERT-like models finetuned on our dataset achieve comparable performance. Furthermore, we show that our dataset enables finetuning of LLMs such as GPT-3.5 that achieve substantially better performance than strong zero-shot baselines.
Our work offers a comprehensive evaluation and a high-quality benchmark on English slang based on the OpenSubtitles corpus, serving both as a publicly accessible resource and a platform for applying tools for informal language processing.\footnote{Code and dataset available at: \url{https://github.com/amazon-science/slang-llm-benchmark}}
\end{abstract}



\section{Introduction}

\begin{figure}[t!]
 \centering
	\begin{subfigure}[b]{0.95\linewidth}
		\includegraphics[width=\linewidth]{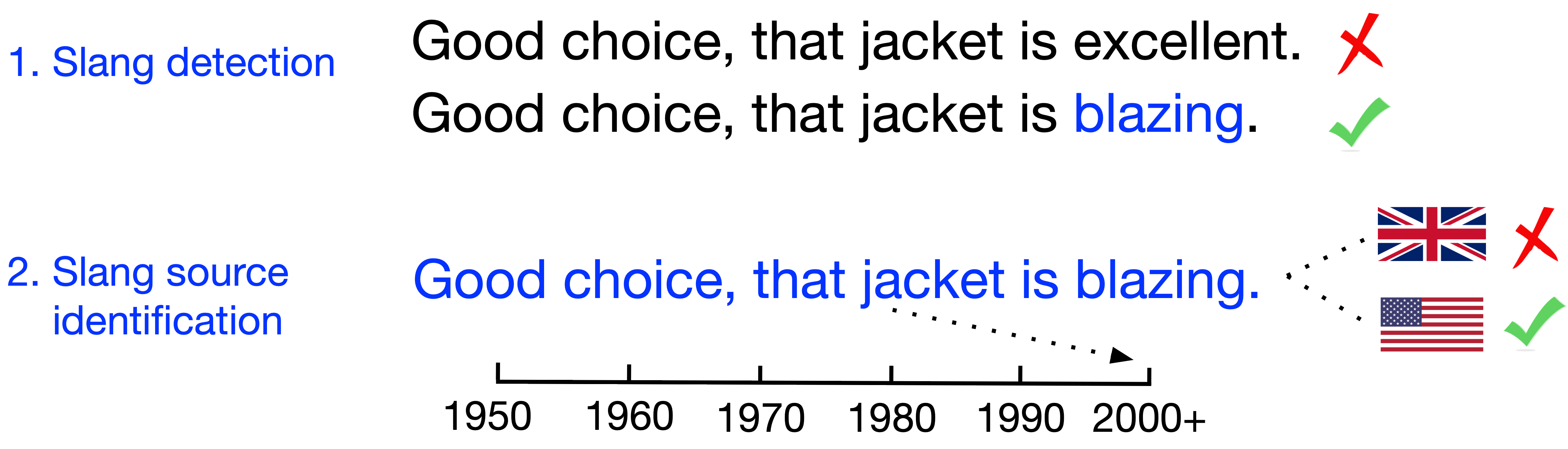}
	\end{subfigure}
	\caption{Overview of tasks used to probe knowledge of slang in LLMs.} 
	\label{fig1}
\end{figure}

Large language models (LLM) are the core engines of widely used applications such as ChatGPT. While the technology is becoming increasingly pervasive, it is important to understand its abilities and limitations with input from diverse forms of language use. Here, we focus on the case of slang - a common type of informal language that is ubiquitous across day-to-day conversations~\cite{mattiello05, eble12}. Figure~\ref{fig1} illustrates the relevance of slang in natural language processing (NLP).
When describing a good jacket, one can make different word choices such as \textit{excellent} and \textit{blazing}. Even though the intended meaning is the same across both word choices, we might expect a significant difference in performance caused by an LLM's lack of knowledge about slang. 
Recent work in computational modeling of slang has suggested that pre-trained LLMs assign much lower probabilities to slang compared to their literal equivalents~\cite{sun21, sun22}, suggesting that models such as BERT~\cite{devlin19} lack knowledge of slang. 

Knowledge of slang in LLMs has important implications beyond automated processing of informal language. This is the case because the use of slang explicitly reflects one's social identity~\cite{labov72, labov06, eble12}. For example, the use of \textit{blazing} to express `Something excellent' emerges from the US whereas it expresses `Anger' in the UK~\cite{green10}. Previous work has shown that the performance of NLP systems can substantially differ across language generated by different demographic groups stratified by age, gender, region, or ethnicity~\cite{hovy15b, hovy16, blodgett17, tatman17, buolamwini18, koeneche20} and can potentially introduce representational harm~\cite{blodgett20}.
Given slang's close ties with social identity, a competent language model may also accurately reveal a slang user's identity. While such information can be used to improve NLP performance~\cite{volkova13, hovy15}, the use of slang may also lead to an increased risk of personal information exposure.

Despite these important implications, LLMs have not been rigorously evaluated across a wide range of models on tasks pertaining to slang. The main challenge lies in the lack of high-quality datasets that are publically accessible. Furthermore, existing dictionary-based data sources (e.g., \citealp{green10}) do not include useful meta-data such as the literal paraphrase of a slang usage. For example, having a pair of sentences as illustrated in Figure~\ref{fig1} where the only difference lies in the slang and its paraphrase (\textit{blazing} and \textit{excellent} respectively) allows us to probe the LLMs in a controlled setting. To address these challenges, we collect a new publically accessible dataset of slang usages based on the OpenSubtitles corpus~\cite{lison16}. Using this dataset, we systematically evaluate the LLMs' knowledge of slang, with a particular focus on the widely adopted GPT models~\cite{brown20, openai23}.\footnote{We focus on GPT models but our evaluative framework can be extended to evaluate other LLMs.} We show that while the LLMs contain considerable knowledge about slang, task-specific finetuning is still essential in achieving state-of-the-art performance.



We focus on two core tasks for informal language processing, illustrated in Figure~\ref{fig1}. First, we evaluate the extent to which LLMs can reliably detect slang usages in natural sentences. Second, we assess whether LLMs can be used to identify regional-historical sources\footnote{Here, the source refers to the extra-linguistic contexts associated with a particular slang usage. Although this would naturally correlate with the origin of a slang (e.g., slang specific to a dialect), our task focuses on predicting the extra-linguistic contexts associated with a particular slang usage and examines whether the models contain knowledge about such usage tendencies.} of slang via a text classification task. Finally, we examine the semantic knowledge of slang in LLMs to understand the differences in their representation of slang versus conventional language use. Throughout our evaluation, we also draw close attention to performance discrepancies based on different demographic variables and discuss their implications toward fairness and privacy.

We make the following contributions in this paper: 1) A dataset containing thousands of human annotated English slang usages in movie subtitles, contributing a novel publically available benchmark of slang for evaluation and finetuning; 2) A rigorous evaluation of large language models' knowledge of slang, including important tasks such as slang detection;
3) A discussion of the implications of such knowledge and how it may affect fairness and privacy in NLP.





\section{Related Work}

\subsection{Deep learning for slang}

Previous work on automatic processing of slang has successfully applied deep learning based techniques to address tasks such as detection~\cite{pei19}, generation~\cite{sun19, sun21}, interpretation~\cite{ni17, sun22}, as well as predicting word formations~\cite{kulkarni18, wibowo21} of slang. 
These tasks are difficult partly due to slang's low resource nature. Our work investigates whether the large scale training of LLMs such as GPT-4 can alleviate this difficulty, and if so, whether GPT's representations reflect semantic knowledge of slang that has been injected in previous methods. We also re-evaluate the slang detection task using modern architectures and contribute the first publically accessible benchmark for slang detection.

Recently, mechanisms underlie both language variation~\cite{lucy21, sun22b} and semantic change~\cite{keidar22} in slang have been extensively studied, with many important features attributed to demographic variables such as age and community membership. We extend this line of work by probing recent large language models for knowledge of slang's demographic source.



\begin{table*}[t!]
    \small
	\centering\makebox[\textwidth]{
		\begin{tabular}{lrrrrrr}
		    
			\multirow{3}{*}{Dataset}&\multicolumn{3}{c}{Slang detection and probing}&\multicolumn{2}{c}{Slang source identification}&\multirow{3}{*}{\shortstack[r]{Publically\\accessible}}
			\\
            \cline{2-6}
            \addlinespace[0.1cm]
            &Slang-containing&Non-slang&Word-level&Community of&Time of&
			\\
            &sentences&sentences&paraphases&emergence&emergence&
			\\
   
			\addlinespace[0.05cm]
			\hline
			\addlinespace[0.1cm]
            Urban Dictionary &\color{blue}{\Checkmark}&\color{red}{\XSolidBrush}&\color{red}{\XSolidBrush}&\color{red}{\XSolidBrush}&\color{red}{\XSolidBrush}&\color{blue}{\Checkmark}\\
            \addlinespace[0.1cm]
			The Online Slang &\color{blue}{\Checkmark}&\color{red}{\XSolidBrush}&\color{blue}{\Checkmark}&\color{red}{\XSolidBrush}&\color{red}{\XSolidBrush}&\color{red}{\XSolidBrush}\\
            Dictionary (OSD) &&&&&& \\
            \addlinespace[0.1cm]
            Green's Dictionary &\color{blue}{\Checkmark}&\color{red}{\XSolidBrush}&\color{red}{\XSolidBrush}&\color{blue}{\Checkmark}&\color{blue}{\Checkmark}&\color{red}{\XSolidBrush}\\
            of Slang (GDoS) &&&&&& \\
            \addlinespace[0.1cm]
            Reddit Glossaries &\color{red}{\XSolidBrush}&\color{red}{\XSolidBrush}&\color{red}{\XSolidBrush}&\color{blue}{\Checkmark}&\color{red}{\XSolidBrush}&\color{blue}{\Checkmark}\\
            \cite{lucy21} &&&&&&\\
            \addlinespace[0.1cm]
            Indonesian Colloquialism &\color{red}{\XSolidBrush}&\color{red}{\XSolidBrush}&\color{blue}{\Checkmark}&\color{red}{\XSolidBrush}&\color{red}{\XSolidBrush}&\color{blue}{\Checkmark}\\
            \cite{wibowo21} &&&&&&\\
            \addlinespace[0.1cm]
            \textbf{OpenSubtitles-Slang}&\color{blue}{\Checkmark}&\color{blue}{\Checkmark}&\color{blue}{\Checkmark}&\color{blue}{\Checkmark}&\color{blue}{\Checkmark}&\color{blue}{\Checkmark}\\
            \textbf{(OpenSub-Slang)} &&&&&&\\

	\end{tabular}}
	\caption{Summary of datasets for slang in NLP and the availability of important features for a comprehensive benchmark. We contribute a new resource (OpenSub-Slang) that captures all desirable features.}
	\label{tabledata}
\end{table*}

\subsection{Probing knowledge in LLMs}

The popularity of deep learning methods in NLP has prompted much work on analyzing the linguistic knowledge learned by neural networks~\cite{belinkov19, rogers20, belinkov22}.
More recent work has probed LLMs on their knowledge of non-standard language such as metaphors~\cite{aghazadeh22, liu22, wicke23} and linguistic anomalies~\cite{li21}. 

Two prominent frameworks have been introduced to operationalize probing. First, the \textit{behavioral probing} method that assesses differences in behavior of a language model given two similar inputs, where a few tokens of interest differ~\citep[e.g.,][]{linzen16}. For example, one can measure the differences in LM scores between alternative words \textit{excellent} and \textit{blazing} given the same context ``Good choice, that jacket is excellent/blazing''. Another widely adopted probing framework involves the training of probing classifiers~\cite{belinkov22} that append a fine-tuned classification layer to the LM. Instead of finetuning the entire model and strive for the highest accuracy, the probing classifiers evaluate knowledge in a model's representations by freezing all pre-trained weights. One such popular probing method is \textit{edge probing}~\cite{tenney19}, in which representations over all tokens in appropriate spans of text are aggregated to predict a label. The resulting accuracy of classification indicates the level of knowledge a model has acquired with respect to the probing task. 

We apply behavioral probing to examine an LLM's confidence in predicting slang usage by comparing LM probabilities of corresponding slang and literal tokens in the same usage context. We apply edge probing in slang detection and slang source identification to analyze an LLM's knowledge of slang's usage and demographic identity.

\section{Data}

\subsection{Limitations of existing resources of slang}


Recent interest in NLP for slang has resulted in a good collection of large-scale datasets for slang. 
Although resources such as the Urban Dictionary are large in scale, the quality of data can be quite poor~\cite{swerdfeger12}. Meanwhile, authoritative sources such as the Green's Dictionary of Slang~\cite{green10} cannot be publically distributed due to copyright restrictions.

The existing datasets are often specified in dictionary format where each entry corresponds to a pair of word and definition sentence. Many datasets include additional features (summarized in Table~\ref{tabledata}) such as the usage context of a slang term (e.g., the sentence `Good choice, that jacket is blazing' is a usage context containing the slang \textit{blazing}), demographic sources such as the community and time of emergence, and word-level literal paraphrases of the slang (e.g., \textit{excellent} is a literal paraphrase of \textit{blazing}). These additional features are often desirable in model evaluation: The usage contexts are important because they allow the slang usages to be embedded in natural sentences; the demographic sources allow us to analyze how regional-historical variation affects performance; finally, literal paraphrases of the slang allow us to test our models against comparable literal baselines.

Previously, \citet{ni17} released a subset of Urban Dictionary data that contains 982,281 entries with associated context sentences. While sentence-level paraphrases of informal language have been collected in previous work~\cite{xu13, dey16, wibowo20, aji21}, few exists at word-level. \citet{wibowo21} collected a set of word-level literal-to-slang paraphrases in Indonesian but no usage context sentences were provided. \citet{sun22} manually annotated a small subset of 102 sentences from the Online Slang Dictionary (OSD) with literal paraphrases of the slang word that fit into the context sentence. 
The existing datasets offer a large pool of examples for training but none captures all desirable features at a sufficient scale. To address this limitation, we contribute a new benchmark dataset of slang usages from movie subtitles that capture all useful features.

\subsection{OpenSub-Slang dataset}

We contribute a new dataset based on movie subtitles from the OpenSubtitles\footnote{\url{http://www.opensubtitles.org/}} corpus~\cite{lison16} that captures all of usage context sentences, demographic information including the region (US or UK) and the year to which the corresponding movie was produced, and word-level literal paraphrases for all slang terms.
We choose to construct a dataset based on OpenSubtitles because movie subtitles contain utterances that better reflect natural conversations, diversifying existing dictionary-based resources containing example usage sentences that are specifically selected to convey the meaning of a slang. Also, metadata associated with the movies allow us to easily obtain demographic information about the slang usages. Finally, the multilingual nature of OpenSubtitles offers potential for multilingual extension in the future, where current NLP research on slang focuses primarily on English.

We sample 100 English movies from the OpenSubtitles corpus partitioned evenly across the regions of US and UK. We annotate randomly sampled sentences on Amazon Mechanical Turk with three annotators per sentence. This results in 7,488 sentences containing slang (3,583 unique terms), of which 2,256 sentences have at least 2/3 annotators agreeing on the exact slang term. Out of the 2,256 sentences, we further annotate them to include definition sentences and literal paraphrases. After manual inspection followed by the removal of nonsensical annotations, we obtain 836 sentences with definitions and paraphrases. Detailed annotation procedures can be found in Appendix~\ref{appdata}.

Alongside the slang containing sentences, we also contribute a set of 17,512 movie subtitle sentences that have been agreed by all annotators to not contain slang. This allows us to build a robust evaluation benchmark for slang detection. Previous evaluations such as \citet{pei19} combine slang containing sentences from slang dictionaries with negative samples heuristically drawn from news corpora. 
This approach, however, may jeopardize sentence-level detection evaluation as the model can rely on dataset-specific features instead of detecting slang.
By having annotated negative sentences from the same data source, we can evaluate slang detection in a more controlled setting where the models can no longer rely on dataset-specific features to make predictions.




\section{Experiments}

\subsection{Models}

We perform all experiments on three BERT-like models: BERT~\cite{devlin19}, RoBERTa~\cite{liu19}, and XLNet~\cite{yang19} using the pre-trained \textit{bert-large-cased}, \textit{roberta-large}, and \textit{xlnet-large-cased} models respectively from the transformers library~\cite{wolf20}. 
We also evaluate a series of GPT models accessed via the OpenAI API, including GPT-3 (\textit{text-davinci-002}), GPT-3.5 (\textit{gpt-3.5-turbo-0613}), and the latest version of GPT-4 (\textit{gpt-4-1106-preview}). Whenever applicable, we also apply finetuning on the same GPT-3.5 model, the newest model to which the authors have finetuning access for.\footnote{Finetuning for GPT-3.5 is completed using a blackbox API provided by OpenAI. Although it is commonly believed that OpenAI does not perturb all model weights during finetuning, the authors do not have direct access to GPT-3.5 to verify the exact training scheme being used.}
For model interpretation, we obtain GPT-3 embeddings using \textit{text-similarity-davinci-001}.

\subsection{Slang detection}

We first ask whether large language models can be used to detect slang's presence in natural sentences. Previous work has found that slang usages have salient features such as Part-of-Speech shifts that are uncommon in literal word usage~\cite{pei19}. A model that encodes knowledge about such characteristics should thus be able to detect slang usages in natural sentences. To evaluate this, we perform edge probing on two slang detection tasks for three BERT-like masked language models: BERT, RoBERTa, and XLNet. In addition, we evaluate the GPT models in both zero-shot and fine-tuned settings. We probe GPT in both a zero-shot setting to evaluate its inherent knowledge and also a fine-tuned variant that has seen the same training examples as the BERT-like models.

\paragraph{Task.}

Given a set of sentences, we evaluate slang detection at both sentence-level and word-level:

\begin{enumerate}[label={(S\arabic*)}, align=left]
    \item Good choice, that jacket is \textbf{blazing}.
    \item Good choice, that jacket is excellent.
\end{enumerate}
In the sentences above, S1 contains a slang usage from the word \textit{blazing} and no slang is used in S2. For \textit{sentence-level detection}, binary classification will be performed to determine whether a slang usage exists within the sentence. For example, S1 containing \textit{blazing} will be a positive example while S2 with \textit{excellent} will be a negative example. In \textit{word-level detection}, we perform a sequence tagging task to identify the specific words that are slang. In the example above, the word \textit{blazing} in S1 should be labeled as slang while all other words in both sentences should have the null label. Detailed experiment setup can be found in Appendix~\ref{appex1}

 	\begin{table}[t!]
		\begin{subfigure}[b]{\linewidth}
			\centering
			\caption{Sentence-Level detection}
			\small
			\begin{tabular}{lrrr}
				Model & P & R & F1 \\
				\hline
				\addlinespace[0.1cm]
				BERT & 80.1 & 83.3 & 81.6 \\
                RoBERTa & 81.3 & 87.5 & 84.2 \\
                XLNet & 67.5 & 64.3 & 64.6 \\
                \addlinespace[0.1cm]
                GPT-3 zero-shot & \textbf{90.0} & 74.4 & 81.4 \\
                GPT-3.5 zero-shot & 87.5 & 80.8 & 84.0 \\
                GPT-4 zero-shot & 88.2 & 80.9 & 84.4 \\
                \addlinespace[0.1cm]
                GPT-3.5 finetuned & 84.3 & \textbf{96.8} & \textbf{90.1} \\
                
			\end{tabular}
		\end{subfigure}\hfill
		
		\vspace{0.2cm}
		
		\begin{subfigure}[b]{\linewidth}
			\centering
			\caption{Word-Level detection}
			\small
			\begin{tabular}{lrrr}
				Model & P & R & F1 \\
				\hline
				\addlinespace[0.1cm]
				BERT & 75.5 & 62.5 & 68.3 \\
                RoBERTa & 74.9 & 68.2 & 71.4 \\
                XLNet & 62.4 & 43.3 & 51.0 \\
                \addlinespace[0.1cm]
                GPT-3 zero-shot & 49.2 & 59.9 & 54.0 \\
                GPT-3.5 zero-shot & 57.6 & 73.2 & 64.5 \\
                GPT-4 zero-shot & 60.4 & 68.2 & 64.1 \\
                \addlinespace[0.1cm]
                GPT-3.5 finetuned & \textbf{74.5} & \textbf{81.3} & \textbf{77.8} \\
                
			\end{tabular}
		\end{subfigure}\hfill
		
		
		\caption{Slang detection results of LLMs shown in precision (P), recall (R), and F1 Scores (F1).}
		\label{tabledet}
	\end{table}

\paragraph{Results.} We evaluate slang detection on sentences from the OpenSubtitles-Slang dataset. Table~\ref{tabledet} shows the results of both sentence-level and word-level slang detection. We observe that for both tasks, BERT and RoBERTa have much better performance than XLNet on slang detection.
While the fine-tuned version of GPT-3.5 performs substantially better than the BERT-like models, the fine-tuned BERT and RoBERTa models can still perform comparably or better than the zero-shot GPT models although having much less parameters. For word-level detection, we observe that the GPT models often have difficulty conforming to sequence labeling instructions without finetuning, resulting in low precision. Overall, we find that GPT models to encode more relevant knowledge that allows the detection of slang's presence but finetuning is nevertheless essential in achieving good performance.

\begin{figure}[t!]
\centering
		\begin{subfigure}[b]{0.95\linewidth}
			\includegraphics[width=\linewidth]{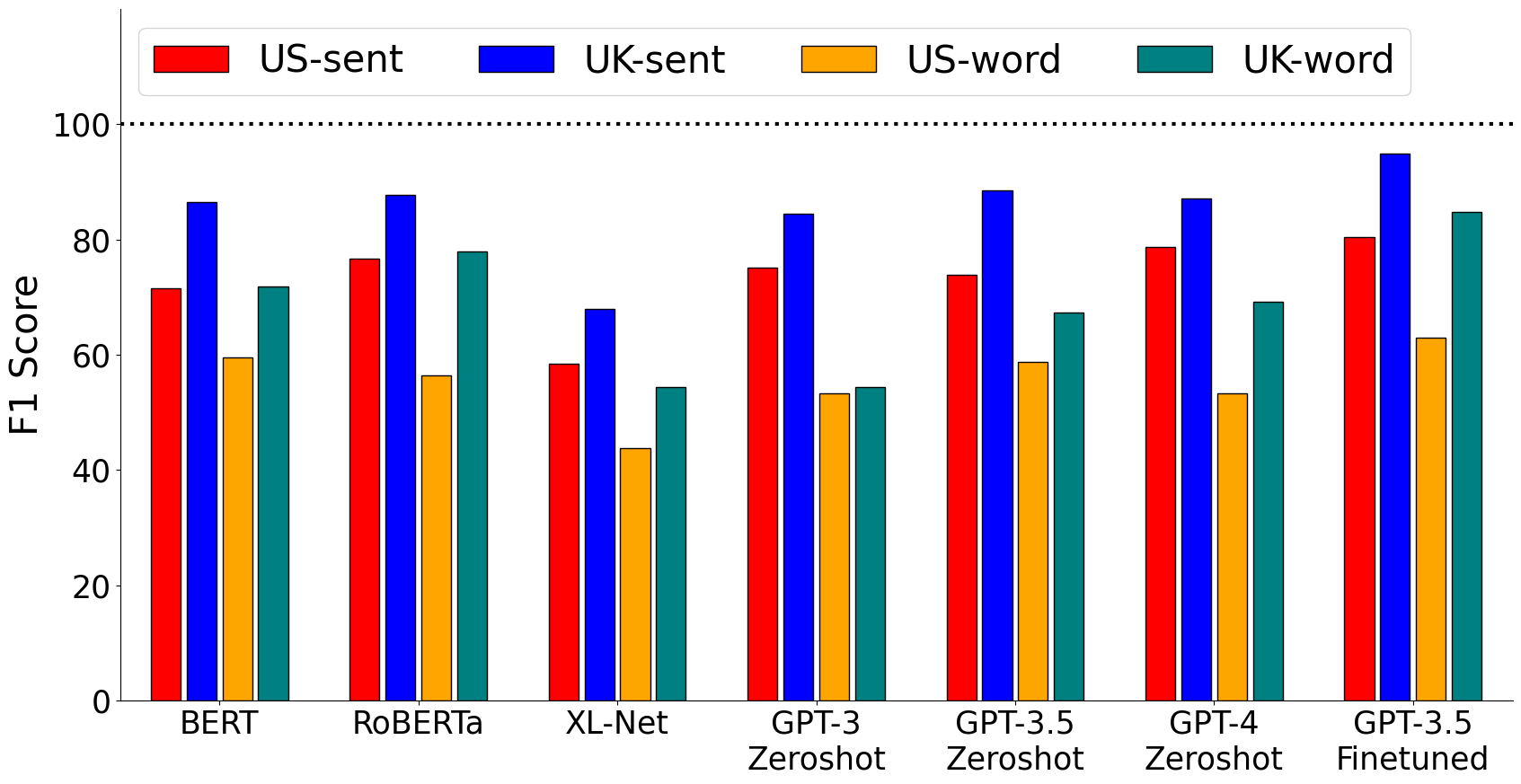}
		\end{subfigure}\
		\caption{Slang detection performance by region.} 
		\label{figdetectreg}
	\end{figure}

\begin{figure}[t!]
\centering
		\begin{subfigure}[b]{0.95\linewidth}
			\includegraphics[width=\linewidth]{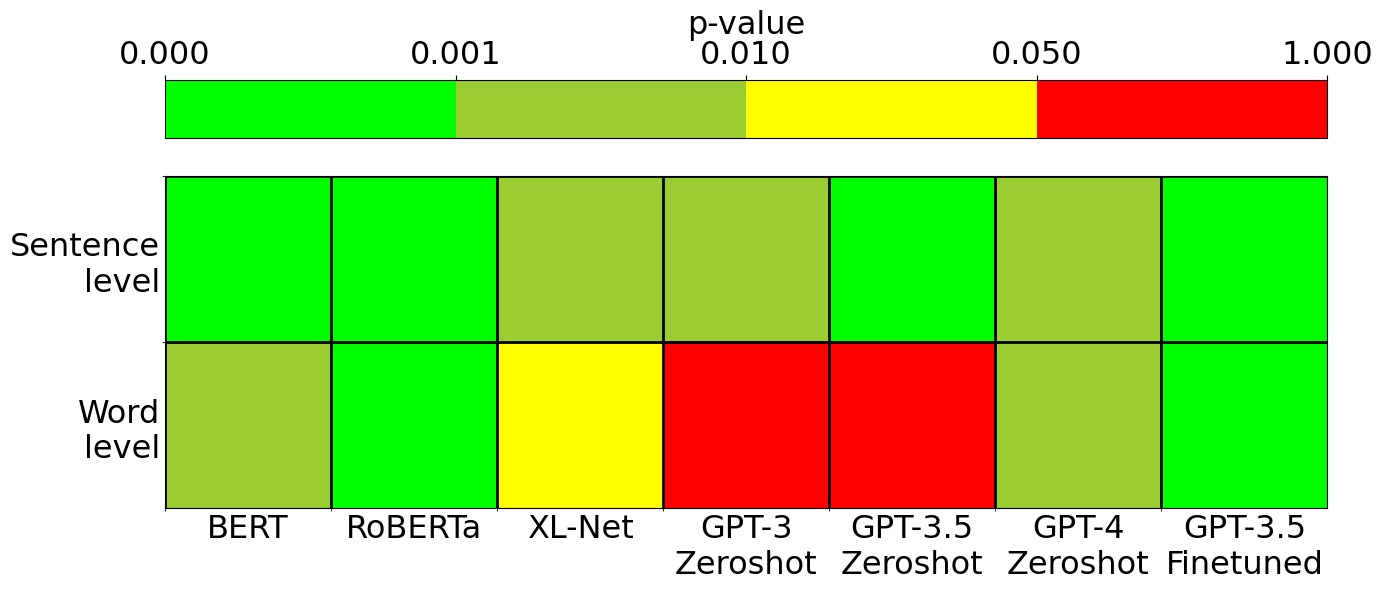}
		\end{subfigure}\
		\caption{Significance level of the regional discrepancies in slang detection performance at both the word-level and the sentence-level. We perform a one-sided test to evaluate whether detection performance on UK slang is indeed significantly better than that of US slang.} 
		\label{figdetectsig}
	\end{figure}


We also partition the test set by region, which yields 234 sentences from the US and 340 sentences from the UK. Figure~\ref{figdetectreg} shows model performance discrepancies between these two regions. We find a consistent trend that slang usages from the UK are being detected more frequently than those from the US, except for the zero-shot GPT-3 models on word-level detection. To account for the small sample size, we perform permutation tests (permuting the region labels) to evaluate the significance of the observed discrepancies. Figure~\ref{figdetectsig} shows the statistical significance of our findings after 20,000 randomized permutation trials for each condition. Overall, we observe that the performance discrepancy in stronger GPT models to be more prominent but not much more than those in smaller BERT-like models.

 \begin{figure}[t!]
 \centering
		\begin{subfigure}[b]{0.95\linewidth}
			\caption{OpenSubtitles-Slang - Region}
			\includegraphics[width=\linewidth]{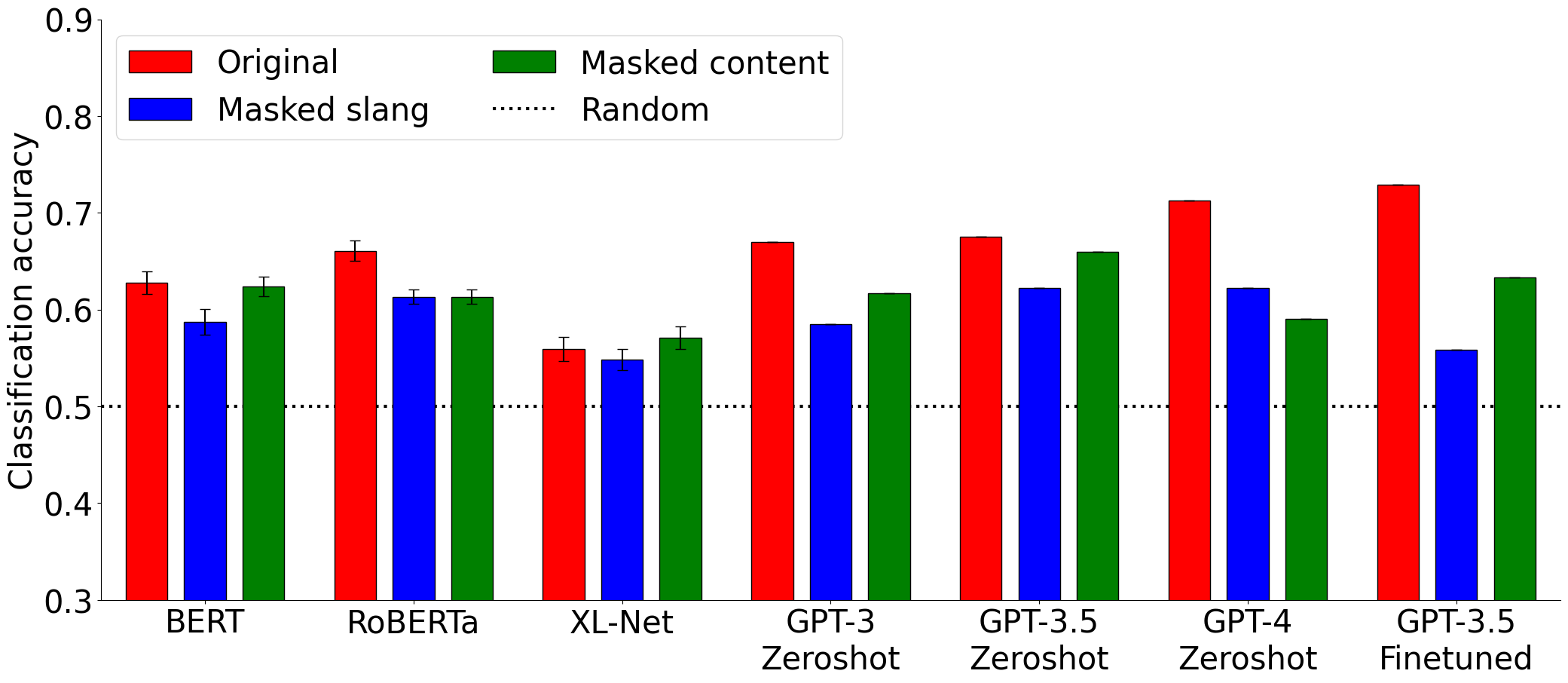}
		\end{subfigure}\
		\begin{subfigure}[b]{0.95\linewidth}
			\vspace{0.2cm}
			\caption{Green's Dictionary of Slang - Region}
			\includegraphics[width=\linewidth]{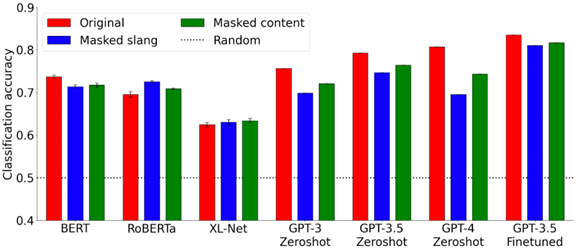}
		\end{subfigure}
        \begin{subfigure}[b]{0.95\linewidth}
			\vspace{0.2cm}
			\caption{Green's Dictionary of Slang - Age}
			\includegraphics[width=\linewidth]{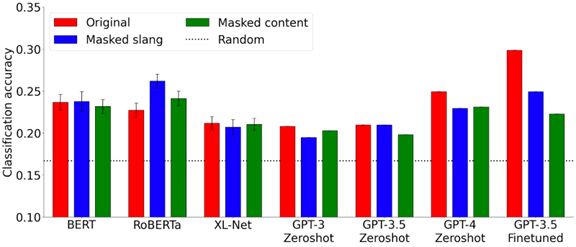}
		\end{subfigure}
		\caption{Classification performance on slang source identification tasks.} 
		\label{figinferreg}
	\end{figure}

\subsection{Slang source identification}

We directly probe large language models' knowledge in identifying a slang's demographic identity. Given that slang is highly reflective of a user's social identity~\cite{labov72, labov06, eble12}, we expect better-performing models to gain such knowledge. We evaluate the extent of such knowledge by probing a text classification task.

\paragraph{Task.} Given a sentence containing a slang usage, we ask the model to classify its source (e.g. region and age). For example, the following sentences should be classified into US as supposed to UK:

\begin{enumerate}[label={(S\arabic*)}, align=left]
    \item Good choice, that jacket is \textbf{blazing}.
    \item Good choice, that jacket is [MASK].
    \item Good [MASK], that jacket is blazing.
\end{enumerate}
We compare the classification performance with sentences containing slang (S1) and corresponding sentences with the slang term masked out (S2). We also include another control task by masking out a random content word in the sentence other than the slang word (S3). For models that use slang as a salient feature to identify demographics, we expect masking out the slang to result in much inferior performance but the performance should not deteriorate as much when masking out a random word.

\paragraph{Results.} Figure~\ref{figinferreg} shows the source identification results on both OpenSubtitles-Slang and Green's Dictionary of Slang for region and age. Overall, we observe that the zero-shot GPT-3 model perform comparably with the BERT-like models and the GPT-4 model is consistently better at predicting demographics compared to earlier models. Although the finetuned GPT-3.5 model achieves the best accuracies across all experiments, the performance is not much better compared to zero-shot GPT-4 when predicting region, whereas finetuning drastically improves the accuracy in age prediction. 
Furthermore, we observe that GPT-3 shows a consistent trend in using slang as a salient feature in predicting demographic identity, indicated by much lower classification accuracies when the slang terms are removed, while the accuracy loss is often not as pronounced when masking out a random word. We also observe this trend in newer generations of GPT models, though it is less pronounced compared to GPT-3.
This behavior is generally not observed in the BERT-like models.


In the BERT-like models, we make two observations suggesting that these
models lack the ability to link slang usages to user demographics. First,
the source identification performance decreases when a slang term is
masked out, and it experiences a similar decrease when a random word is
masked out. This observation suggests that decrease in model performance is tied to removal of words but not specifically to slang usages. Second, the source
identification performance improves after a slang term is removed. Here,
it is plausible that the BERT-like models do not faithfully capture slang
meaning. With slang terms removed, the models may become more certain
about the underlying meaning conveyed by a sentence. In both cases, we do
not find conclusive evidence that the BERT-like models are able to make
connections between slang and the source of a sentence.

\subsection{Model interpretation} \label{ex2}

We perform interpretive analysis to examine whether large language models have gained structural semantic knowledge about slang through large scale training. 
We do so by first comparing the usage probabilities of slang and their corresponding literal paraphrase tokens.
Here, high model probabilities on slang tokens reflect a model's confidence in predicting the slang term to be used within the specified linguistic context, thus having good distributional semantic knowledge of a slang's meaning. We also analyze sentence embeddings generated by the LLMs on conventional and slang dictionary senses to examine whether geometry of the underlying representation space reflects structural semantic knowledge of slang.


\paragraph{Task.} Given a sentence containing slang, we examine a model's predictive confidence in a slang usage by measuring the LM probability associated with the slang word. If a literal paraphrase of the word is available, we compare the probability of the slang word with its literal counterpart:

\begin{enumerate}[label={(S\arabic*)}, align=left]
    \item Good choice, that jacket is \textbf{blazing}.
    \item Good choice, that jacket is \textbf{excellent}.
\end{enumerate}
For the example sentences above, we measure language model probabilities assigned to both the slang word \textit{blazing} and the literal word \textit{excellent} given the exact preceding context. Detailed experiment setup can be found in Appendix~\ref{appex2}

\paragraph{Metrics.}

We report two metrics to compare an LLM's predictive confidence in slang usages relative to their literal counterparts. Let $\mathcal{S}_i$ denote the language model probability assigned to the slang word in the $i$'th sentence and similarly $\mathcal{L}_i$ for the literal word's probability. The mean ratio compares the aggregate probability mass assigned to each word type over a sample of sentences:
\begin{align}
    r_{mean} = \frac{\sum_{i} \mathcal{S}_i}{\sum_{i} \mathcal{L}_i}
\end{align}
Here, we aggregate over probabilities for each type instead of individual ratios to avoid over-emphasizing outlier slang that the model is either very confident or very impoverished on. For individual ratios between two word types, we report the median ratio to downplay the effect of outliers. A value above 1 means that more slang words have higher probabilities than their literal paraphrases:
\begin{align}
    r_{median} = \mathrm{median}_i\frac{\mathcal{S}_i}{\mathcal{L}_i}
\end{align}

\begin{figure}[t!]
\centering
	\begin{subfigure}[b]{0.99\linewidth}
		\includegraphics[width=\linewidth]{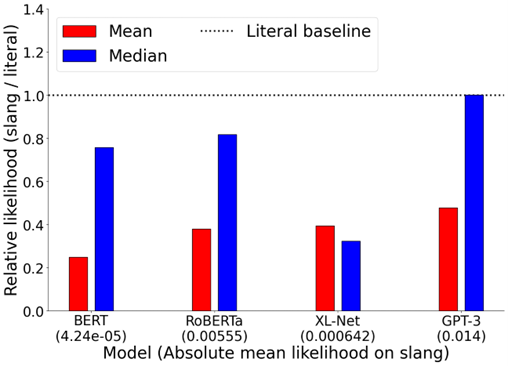}
	\end{subfigure}\vfill
	\caption{Likelihood ratios between samples of corresponding slang and literal tokens.} 
	\label{figmlm}
\end{figure}



\paragraph{Results.} Figure~\ref{figmlm} summarizes the results for sentences from OpenSubtitles-Slang.
We observe that for all of BERT, RoBERTa, and GPT-3,\footnote{We only perform analysis on GPT-3 because OpenAI no longer provides token probabilities (on prompted words) and embeddings for newer generation GPT models.} the models have much higher median ratio than mean ratios, suggesting that these models are confident on many of the slang terms in the dataset but impoverished on a select subset with much higher probabilities assigned to the paraphrases. In absolute terms, GPT-3 also assigns much higher probability scores to slang terms compared to the BERT-like models.

\begin{figure}[t!]
\centering
	\begin{subfigure}[b]{0.99\linewidth}
		\includegraphics[width=\linewidth]{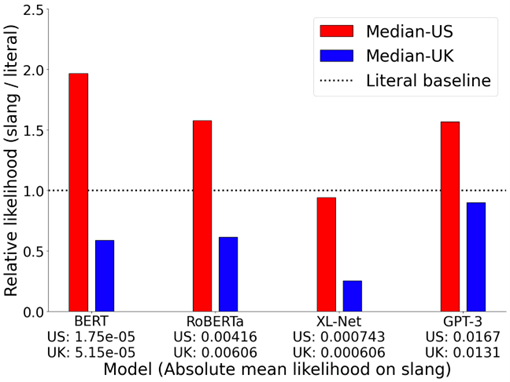}
	\end{subfigure}\vfill
	\caption{Median ratios across sentences from different regions.} 
	\label{figmlmreg}
\end{figure}

\begin{table}[t!]
		\begin{subfigure}[b]{\linewidth}
			\centering
			\small
			\begin{tabular}{lrrr}
				Model & OSD & GDoS & UD \\
				\hline
				\addlinespace[0.1cm]
				fastText & 0.35 $\pm$ 0.033 & 0.30 $\pm$ 0.010 & 0.31 $\pm$ 0.037 \\
                SBERT & 0.32 $\pm$ 0.033 & 0.32 $\pm$ 0.010 & 0.28 $\pm$ 0.034 \\
                \addlinespace[0.1cm]
                GPT-3 & 0.31 $\pm$ 0.032 & 0.31 $\pm$ 0.011 & 0.30 $\pm$ 0.035
                
			\end{tabular}
		\end{subfigure}\hfill
		
		
		\caption{Normalized ranks (between 0 and 1, lower is better) of a word's slang definition embedding towards its conventional definition embedding over entries in The Online Slang Dictionary (OSD), Green's Dictionary of Slang (GDoS) and Urban Dictionary (UD). We compare the embeddings produced by GPT-3 against those computed in \citet{sun21} using fastText~\cite{bojanowski17} and Sentence-BERT~\citep[SBERT;][]{reimers19}.}
		\label{tableembed}
	\end{table}

 \begin{table*}[t!]
    \small
	\makebox[\textwidth]{
		\begin{tabular}{lrrrrrrrr}

            &\multicolumn{4}{l}{Example 1} & \multicolumn{4}{l}{Example 2}\\
			\hline
			\addlinespace[0.1cm]
   
            Sentences&\multicolumn{4}{l}{* We can't keep doing this sh$\star$t, Charlie. * }&\multicolumn{4}{l}{* Knock it off. *}\\
            &\multicolumn{4}{l}{Look, I don't know what I said to you in there that}&\multicolumn{4}{l}{You'd \textit{kill} to be in his place.}
            \\
            &\multicolumn{4}{l}{got you so \textit{pissed} off but I'm sorry, Charlie, all right?}& \\
            &\multicolumn{4}{l}{* All right. *}&\multicolumn{4}{l}{* - Okay b$\star$tch, I'm ready. *}\\

            \addlinespace[0.1cm]

            Slang & \multicolumn{4}{l}{pissed} & \multicolumn{4}{l}{kill}\\
            Literal paraphrase & \multicolumn{4}{l}{angry} & \multicolumn{4}{l}{agree}\\
            Definition sentence & \multicolumn{4}{l}{Annoyed; anger.} & \multicolumn{4}{l}{To agree with someone or about something.} \\
            Region & \multicolumn{4}{l}{US} & \multicolumn{4}{l}{US}\\

            \addlinespace[0.15cm]

            \underline{Model scores} &\underline{BERT} & \underline{RoBERTa} & \underline{GPT-3} && \underline{BERT} & \underline{RoBERTa} & \underline{GPT-3} \\

            \addlinespace[0.05cm]

            Detection & 0.697 & 0.811 & 0.393 && 0.644 & 0.134 & 0.621\\


            Source identification & 0.491 & 0.436 & 0.762 && 0.873 & 0.710 & 0.879\\


            Model confidence & 0.035 & 0.057 & 1.567 && 9.654 & 0.170 & 3.120\\
	    \end{tabular}}
     
	\caption{Example entries and their corresponding model scores from BERT, RoBERTa, and zero-shot GPT-3 respectively. Asterisks indicate extra context sentences not seen by the model.}
	\label{tableex}
\end{table*}

Next, we examine performance discrepancy by partitioning the data based on its region. This results in 59 sentence pairs from the US and 161 sentence pairs from the UK. Results from Figure~\ref{figmlmreg} show that all models evaluated are much more confident in generating US slang compared to UK slang. GPT-3, however, has substantially less discrepancy in performance between the two regions due to it being more confident in UK slang. We also measure absolute probabilities assigned to slang tokens in context sentences extracted from Green's Dictionary of Slang entries. By stratifying across different age groups and regions, we observe that the systems are much less confident on contemporary slang and only within this group that UK slang tends to receive much lower scores than US slang. Details results can be found in Appendix~\ref{appgreens}.

Interestingly, we observe a reverse trend in discrepancy compared to the case in slang detection. Specifically, being less confident, in terms of probability, on UK slang terms makes it easier for the models to detect them. 
Indeed, we observe that US slang terms are often assigned higher probability scores than their literal counterparts, suggesting that slang usages from the US have been seen more frequently in the training data and the models use frequency as a salient feature to characterize slang.

\paragraph{Analysis.} We look at text embeddings produced by GPT-3 to examine whether they encode semantic knowledge of slang. We adopt the benchmark proposed by \citet{sun21} that compares sentence embeddings of definition sentences. In this evaluation, the embedding of a slang definition is taken as an anchor and its semantic distances toward conventional definitions of words are computed. The distances are then ranked among all words in the lexicon and we expect the groundtruth word to receive a good rank. As an example, for \textit{blazing} that can be used as slang to express `something excellent', we expect the slang definition to be semantically close to the conventional definition of \textit{blazing} - `burning brightly' compared to definitions of other words in the lexicon. \citet{sun21, sun22} showed that this metric reflects the semantic knowledge of slang encoded in a model and is directly tied to performance in slang generation and interpretation. Table~\ref{tableembed} shows the results of this evaluation. While GPT-3 shows better performance on slang than the BERT-like models on extrinsic tasks, we do not observe any significant difference in the underlying geometry of the representations. This gives further evidence that GPT-3's source of knowledge comes from frequent instances of slang usage seen during training and simply treats them as additional ``conventional'' senses. It has yet been able to (or decided not to) encode any structural knowledge of slang into its representations.

\paragraph{Examples.} We find two entries with definition and paraphrase annotation that appear in the test set for both slang detection and source identification. For each example, we show the respective model performances in Table~\ref{tableex}. We show results for the best performing BERT-like model BERT and RoBERTA, as well as the zero-shot version of GPT-3 where probability scores are available. For the classification based tasks, we report each model's confidence on the true label (i.e. $P($True label$)$). For model confidence, we report the ratio between the LM scores of the slang word and its literal paraphrase (i.e. $\mathcal{S}_i/\mathcal{L}_i$). For BERT-likes, we report the normalized probabilities from the final classification layer. For GPT-3, we use the top-5 probabilities assigned to the response token by the OpenAI API. We then sum and normalize all token probabilities that correspond to one of the classes.

We observe that although GPT-3 reliably identifies and assigns high probabilities to both slang usages, it still failed to detect the slang \textit{pissed} in Example 1. We find this trend to be consistent for slang detection test examples that have paraphrase annotations (32 examples) where negative correlations exist between model confidence scores and detection probability for all of BERT ($r=-0.433$), RoBERTa ($r=-0.458$), and GPT-3 ($r=-0.220$). This is consistent with our earlier finding where all models tend to consider less frequent usages as slang. We perform a similar experiment on source identification examples (21 examples) and find the correlations to be much weaker (BERT: $r=0.020$, RoBERTa: $r=-0.121$, GPT-3: $r=0.276$), although GPT-3 tend to better identify a slang's region when it has high confidence.

\section{Conclusion}

We offered a comprehensive investigation of slang knowledge in large language models. We show that larger GPT models are more knowledgeable about slang compared to BERT-like models in 1) better detecting slang in natural sentences, 2) more accurately identifying the regional source and time period of slang usages, and 3) better predicting slang usages relative to their literal counterparts. Despite the superiority of GPT in these slang processing tasks, we did not find evidence that it represents or encodes slang as a special form of language. It is conceivable that GPT has learned to process slang by treating slang usages as rare meanings of words expressed in appropriate linguistic contexts.

In the identification of region and age of slang usages, we observed that all models tend to perform poorly on slang from the UK (compared to US slang) and more contemporary slang (compared to historical slang), likely due to impoverished training data. However, we found that GPT models are no more biased compared to earlier BERT-based models and that it shows comparable discrepancy in processing slang across regions. Additionally, we observed that GPT models contain good knowledge about the demographic identities of a slang usage in context. This capability may have implications for privacy in many scenarios (e.g., automatic data annotation), and users should be aware of the increased risk of identity exposure when using slang in LLM-based applications.

We have provided the first comprehensive probing analysis of large language models on knowledge of slang and have contributed an open benchmark dataset to facilitate future efforts in evaluating and improving large language models on informal language processing.

\section*{Limitations}

In our work, the sets of comparable experiments we can perform have been limited by lack of direct access to GPT models. Finetuning of GPT-3.5, for example, is completed using a blackbox API provided by OpenAI while newer models such as GPT-4 are not available for finetuning. Although it is commonly believed that OpenAI does not perturb all model weights during finetuning, the authors do not have direct access to GPT-3.5 to verify the exact training scheme being used. This may cause inconsistency in the experiment setup involving finetuned models. Also, the lack of access to internal layers of GPT hinders the comparison of intermediate representations in LLMs. For example, we can only analyze probability values from GPT-3 as OpenAI no longer provides access to those values in newer generation models like GPT-3.5 and GPT-4.
Finally, the auto-regressive nature of GPT necessitates the comparison with the BERT-like masked language models in an auto-regressive setup. Although approaches such as \citet{donahue20} have been proposed to enhance GPT-2 to consider bidirectional context, we cannot apply such methods to GPT given the limited access.

We also acknowledge that our work is limited to studying slang in English and is restricted to specific demographic stratum (region and age). We hope that the evaluation framework proposed in this work would enable future work to extend the evaluation towards more varied demographics and languages. We selected OpenSubtitles to build our dataset because of its 
potential in extending the existing evaluation into a multilingual benchmark.



\section*{Ethics Statement}
We acknowledge that many slang-containing sentences annotated contain profanity, sexual references, and/or stereotypical views towards specific groups of our community. Discretion is advised when using the collected datasets. During annotation, we begin our HIT with a disclaimer informing annotators that ``this HIT contains language use that may be offensive or upsetting.''. If the annotator does not provide consent in annotating such language, they may exit the HIT without penalty.
All potentially offensive sentences shown in the example sections of this paper were taken verbatim from the original data source and do not reflect opinions of the authors and their affiliated organizations. The manuscript has been reviewed and approved by an internal ethics committee before submission.

We compensate all human annotators via Amazon Mechnical Turk, regardless of whether the annotated entries were kept after quality control. We compensate all annotators \$0.10 USD for identifying slang in up to 10 sentences and \$0.40 USD for defining and paraphrasing slang in up to 10 sentences. We run all experiments for BERT-like models (BERT, RoBERTa, and XLNet) using an in-house GPU server with 1 Nvidia Titan V GPU and 12 GB of VRAM available to the authors. All GPT model experiments are executed via OpenAI's official API and cost \$77.22 USD in API credits.

We have written permissions to use both The Online Slang Dictionary and Green's Dictionary of Slang for personal research use from the respective authors. We obtained OpenSubtitles data from the Open Parallel Corpus~\citep[OPUS; ][]{tiedemann12} containing user generated movie subtitles. We are not aware of any existing license posted for this dataset but follow all citation requests outlined by its authors.

\section*{Acknowledgements}

We thank the anonymous ARR reviewers and chairs for their constructive comments and suggestions. We thank Jwala Dhamala for feedback on the data annotation process. This work was supported by a QEII-GSST award to ZS and an Amazon Research Award to YX.

\bibliography{custom}

\newpage

\appendix

\begin{table*}[t!]
    \small
	\centering\makebox[\textwidth]{
		\begin{tabular}{lrrrr}
		    
			OpenSubtitles ID&Year&Region&Genres\\
			\addlinespace[0.05cm]
			\hline
            \addlinespace[0.1cm]
			
54446 & 2000 & USA & Adventure, Comedy, Drama \\
135737 & 2000 & USA & Action, Crime, Thriller \\
145382 & 2000 & USA & Drama, Romance \\
185218 & 2001 & USA & Crime, Drama, Romance \\
186160 & 2004 & USA & Comedy, Sport \\
241730 & 2005 & USA & Comedy, Drama \\
3151540 & 2007 & USA & Drama \\
3279503 & 2008 & USA & Crime, Mystery, Thriller \\
3372842 & 2000 & USA & Action, Adventure, Drama \\
3468388 & 2007 & USA & Crime, Drama \\
3546395 & 2009 & USA & Drama \\
3558591 & 2005 & USA & Comedy, Romance \\
3562517 & 2009 & USA & Comedy, Fantasy, Romance \\
3618044 & 2009 & USA & Comedy, Drama \\
3692182 & 2009 & USA & Crime, Drama, Thriller \\
3877824 & 2009 & USA & Horror, Thriller \\
3967329 & 2010 & USA & Drama \\
4109374 & 2010 & USA & Comedy, Drama, Romance \\
4185464 & 2011 & USA & Crime, Drama \\
4218973 & 2011 & USA & Crime, Drama, Horror \\
4473014 & 2011 & USA & Drama, Mystery, Romance \\
4574956 & 2011 & USA & Comedy \\
4728198 & 2001 & USA & Drama \\
4744540 & 2012 & USA & Drama, Sport \\
4953583 & 2013 & USA & Action, Crime, Thriller \\
5036434 & 2012 & USA & Drama \\
5166024 & 2013 & USA & Adventure, Comedy, Drama \\
5178727 & 2010 & USA & Comedy, Drama \\
5340423 & 2013 & USA & Comedy, Drama, Romance \\
5450161 & 2013 & USA & Comedy, Drama, Romance \\
5536320 & 2014 & USA & Biography, Crime, Drama \\
5653079 & 2012 & USA & Comedy, Drama, Romance \\
5697912 & 2012 & USA & Comedy, Drama, Romance \\
5791518 & 2014 & USA & Comedy, Romance \\
5836657 & 2014 & USA & Comedy, Romance \\
5838045 & 2014 & USA & Sci-Fi, Thriller \\
5860680 & 2014 & USA & Drama, Romance, Sci-Fi \\
5891414 & 2014 & USA & Action, Crime, Thriller \\
5905224 & 2012 & USA & Comedy, Romance \\
5922900 & 2012 & USA & Comedy, Horror, Sci-Fi \\
5974299 & 2014 & USA & Comedy, Drama, Romance \\
5987878 & 2006 & USA & Comedy, Romance \\
6173232 & 2014 & USA & Documentary, Music, Sport \\
6185084 & 2015 & USA & Comedy \\
6249260 & 2014 & USA & Comedy, Musical \\
6377252 & 2009 & USA & Action, Crime, Thriller \\
6406429 & 2001 & USA & Drama, Music \\
6441036 & 2013 & USA & Drama, Family, Fantasy \\
6692456 & 2016 & USA & Crime, Drama, Mystery \\
6801883 & 2014 & USA & Crime, Drama, Mystery \\

	\end{tabular}}
	\caption{Meta-data for all US movies used in constructing OpenSubtitles-Slang.}
	\label{tablemovie}
\end{table*}

\begin{table*}[t!]
    \small
	\centering\makebox[\textwidth]{
		\begin{tabular}{lrrrr}
		    
			OpenSubtitles ID&Year&Region&Genres\\
			\addlinespace[0.05cm]
			\hline
            \addlinespace[0.1cm]
			
3120452 & 2006 & UK & Comedy, Drama, Romance \\
3121411 & 2006 & UK & Crime, Drama, Thriller \\
3179568 & 2006 & UK & Crime, Drama \\
3320486 & 2008 & UK & Comedy, Drama, Romance \\
3345059 & 2008 & UK & Crime, Drama \\
3357285 & 2007 & UK & Drama, Romance \\
3472293 & 2009 & UK & Crime, Drama, Mystery \\
3552835 & 2008 & UK & Crime, Drama, Horror \\
3564173 & 2008 & UK & Drama \\
3666051 & 2009 & UK & Action, Crime, Drama \\
3670999 & 2010 & UK & Biography, Drama, Music \\
3807079 & 2005 & UK & Comedy, Crime \\
4030209 & 2003 & UK & Documentary, Music \\
4107485 & 2010 & UK & Comedy, Thriller \\
4136037 & 2010 & UK & Biography, Documentary, Drama \\
4177060 & 2009 & UK & Action, Crime, Drama \\
4204063 & 2009 & UK & Comedy, Drama, Romance \\
4259257 & 2010 & UK & Comedy, Drama \\
4398890 & 2011 & UK & Action, Thriller \\
4471635 & 2010 & UK & Crime, Drama, Thriller \\
4527521 & 2012 & UK & Crime, Thriller \\
4629499 & 2012 & UK & Crime \\
4640913 & 2011 & UK & Comedy, Drama, Music \\
4683078 & 2012 & UK & Drama, Sport \\
4864547 & 2012 & UK & Crime, Drama, Mystery \\
4938516 & 2009 & UK & Mystery, Thriller \\
4987950 & 2011 & UK & Drama \\
5052284 & 2002 & UK & Crime, Drama, Mystery \\
5145968 & 2012 & UK & Horror, Mystery \\
5151994 & 2008 & UK & Action, Biography, Crime \\
5167828 & 2001 & UK & Drama, Mystery, Thriller \\
5204705 & 2012 & UK & Crime, Drama, Thriller \\
5461631 & 2003 & UK & Comedy, Drama, Romance \\
5510712 & 2013 & UK & Action \\
5623414 & 2013 & UK & Comedy, Music \\
5681039 & 2004 & UK & Comedy, Crime, Drama \\
5742017 & 2010 & UK & Action, Crime, Drama \\
5778643 & 2013 & UK & Documentary, Sport \\
5814259 & 2014 & UK & Drama, Musical, Romance \\
5837569 & 2002 & UK & Horror, Thriller \\
6010762 & 2012 & UK & Crime, Drama \\
6107374 & 2010 & UK & Comedy, Drama \\
6224678 & 2014 & UK & Thriller \\
6237485 & 2014 & UK & Drama \\
6244263 & 2014 & UK & Thriller \\
6338678 & 2008 & UK & Drama, Romance, Thriller \\
6782316 & 2009 & UK & Biography, Drama, Sport \\
6910409 & 2014 & UK & Comedy, Drama \\
6997754 & 2012 & UK & Action, Crime, Drama \\
7039857 & 2016 & UK & Comedy \\

	\end{tabular}}
	\caption{Meta-data for all UK movies used in constructing OpenSubtitles-Slang.}
	\label{tablemovie2}
\end{table*}

\begin{table*}[t!]
    \small
	\makebox[\textwidth]{
		\begin{tabular}{ll}
            [Positive Examples]\\
            \addlinespace[0.1cm]
            
            Example 1 & * We're hitting pause after this. *
            \\
            \addlinespace[0.05cm]
            &We get \textit{pinched}, remember whose idea this was, okay?
            \\
            \addlinespace[0.05cm]
            &* Be ready on Friday. *
            \\
            \addlinespace[0.15cm]

            Example 2 & * Now, if it were up to me and they gave me two minutes and a wet towel I would personally \\&tasphyxiate his half-wit so we could string you up on a federal M1 and end this story with a bag \\&on your head and a paralyzing agent running through your veins. *
            \\
            \addlinespace[0.05cm]
            &This isn't f$\star$cking \textit{townie} hopscotch anymore, Doug.
            \\
            \addlinespace[0.05cm]
            &* Be ready on Friday. *
            \\
            \addlinespace[0.15cm]

            Example 3 & * I can't do any more time, Dougy. *
            \\
            \addlinespace[0.05cm]
            &So if we get \textit{jammed} up we're holding court on the street.
            \\
            \addlinespace[0.05cm]
            &* [KNOCKING] *
            \\
            \addlinespace[0.15cm]

            Example 4 & * She really loves you, I can tell. *
            \\
            \addlinespace[0.05cm]
            &Good news for you is you have an \textit{alibi} for the Cambridge job.
            \\
            \addlinespace[0.05cm]
            &* The good news for me is I bet you know something about it. *
            \\
            \addlinespace[0.15cm]

            Example 5 & * What do you call a guy who grows up with a group of people, gets to know their secrets \\&because they trust him, and then turns around and use those secrets against them, put those \\&people in prison?
 *
            \\
            \addlinespace[0.05cm]
            &You'd call him a \textit{rat}, right?
            \\
            \addlinespace[0.05cm]
            &* You know what I call him? *
            \\
            \addlinespace[0.15cm]
            \addlinespace[0.15cm]

            [Negative Examples]\\
            \addlinespace[0.1cm]
            
            Example 1 & * Any clues? *
            \\
            \addlinespace[0.05cm]
            &Any \textit{leads}?
            \\
            \addlinespace[0.05cm]
            &* Anything like that? *
            \\
            \addlinespace[0.15cm]

            Example 2 & * With assault rifles. *
            \\
            \addlinespace[0.05cm]
            &You f$\star$cking \textit{dummies} shot a guard.
            \\
            \addlinespace[0.05cm]
            &* Now you're like a half-off sale at Big and Tall. *
            \\
            \addlinespace[0.15cm]

            Example 3 & * Coughlin, Kristina. *
            \\
            \addlinespace[0.05cm]
            &She had a \textit{kid} with her.
            \\
            \addlinespace[0.05cm]
            &* The mother's at Mass General. *
            \\
            \addlinespace[0.15cm]

            Example 4 & * Do me a favor. *
            \\
            \addlinespace[0.05cm]
            &The weight of this thing \textit{pack a parachute} at least.
            \\
            \addlinespace[0.05cm]
            &* You know the funniest thing about being in prison? *
            \\
            \addlinespace[0.15cm]

            Example 5 & * [SHYNE CRYING] *
            \\
            \addlinespace[0.05cm]
            &I know you'd rather see a \textit{rope around my neck}!
            \\
            \addlinespace[0.05cm]
            &* You're getting the f$\star$ck out of here. *
            \\
            \addlinespace[0.15cm]
	    \end{tabular}}
	\caption{Positive and negative annotations examples shown to annotators prior to annotation. Candidate slang terms are italicized. Sentences marked by asterisks indicate extra context sentences that the annotators are asked to consider but not to make annotations on.}
	\label{tableannoex}
\end{table*}

\section{Data Collection Procedures} \label{appdata}

We sample 100 English movies from the OpenSubtitles corpus for an even distribution across the regions of US and UK, where we identify the region by querying a movie's region of production on IMDb\footnote{\url{https://www.imdb.com/}}. For each region, we randomly shuffle the list of corresponding movies represented in the OpenSubtitles corpus that are produced after the year 2000 and iterate through the list until we have 50 movies. For each movie, the authors manually inspect the corresponding IMDb meta-data to ensure that the movie has a realistic setting in the appropriate region (i.e. US or UK) and that the plot is set after year 1980 to avoid antiquated slang. In addition, we ensure that each selected movie would have sufficient sentences by filtering out all movies with less than 500 subtitle lines. As a result, the most common genre tags are drama, comedy, crime and romance. A complete list of movie meta-data can be found in Table~\ref{tablemovie} and Table~\ref{tablemovie2}.

For each movie, we randomly sample 250 sentences for annotation on Amazon Mechanical Turk\footnote{We opted for random sampling instead of first detecting slang using language models as it would introduce a selection bias to our evaluation}. We restrict the set of annotators to native English speakers who live in the corresponding region of the movie (i.e. US or UK). We first ask annotators to detect sentences containing slang usage and identify the exact slang terms. To define what is considered slang, we provide all annotators with 5 positive examples containing slang and 5 negative examples that closely resemble slang usage. Table~\ref{tableannoex} shows these examples. We obtain these examples from a small scale pilot study and manually verify that all positive examples have exact definition matches in Green's Dictionary of Slang while all slang-like words in the negative sentences do not have corresponding entries in the dictionary. For each annotation, the preceding and succeeding sentences in the movie scripts are also shown to the annotators for contextual awareness but they are only asked to find slang in the main sentence. For each sentence, we ask 3 annotators to perform the same task. Overall, 7,488 sentences are flagged by at least one annotator as containing slang (3,583 unique terms), with 1,844 and 412 sentences flagged by two or all three annotators respectively. We adopt a majority vote scheme and only consider sentences with at least 2 annotators agreeing but include all sentences and annotator confidence scores in the dataset for future users. 

For the 885 sentences with at least 2/3 annotator agreement, we further annotate these sentences by asking one additional annotator to give a definition sentence and a literal paraphrase of the slang. The annotators were directed to both Green's Dictionary of Slang and Urban Dictionary for reference, in this order of preference, and asked to cite a URL for the definition if possible. We manually inspect the annotator responses and remove all that are nonsensical (e.g. writing down the same definition sentence for all slang in a batch). After removing such responses, we obtain 836 sentences with 366 unique slang terms that have both a definition sentence and a literal paraphrase.

During annotation, the annotators were not told about the specific year of release and were asked to annotate the slang solely based on its usage and the corresponding references from slang dictionaries. We do not specify the year during annotation because annotators will likely not have the necessary expertise to differentiate multiple meanings of slang across several time periods.

\section{Experiment Setup} \label{appex}

\subsection{Probing classifiers}

We implement BERT, RoBERTa, and XLNet classifiers using the transformers library~\cite{wolf20} released by Huggingface. For each model, we use the corresponding sequence classification classes for sentence-level detection and source identification, and token classification classes for word-level detection. For all models, we only train weights for the classification layers that are not part of pre-training, except for BERT where we re-train weights for the final pooling layer. We do this to ensure consistency across all models because only BERT has a pre-trained pooling layer used for its next-sentence prediction objective while such a layer does not exist in pre-trained RoBERTa and XLNet. we train each model for 10 epochs and repeat the experiment 20 times with different random initializations. We use Adam~\cite{kingma15} with a learning rate of 0.001. Parameters from the training epoch with the highest validation performance is saved for testing.

For GPT-3.5 finetuning, We train each model once using the same set of training and validation data used to train the BERT-like models and train the model for four epochs using OpenAI's API interface using default parameters with a batch size of 20.

\subsection{Slang detection} \label{appex1}

We use entries from the OpenSub-Slang dataset with an annotator confidence score of 2 or above for positive examples. We use all sentences in which exact one copy of the exact slang identified by the annotators can be found. After filtering, we have 1,913 slang containing sentences. From the set of 17,512 movie subtitle sentences where all 3 annotators labeled as not containing slang, we randomly sample 1,913 sentences to construct a balanced sample. We split the data into 80, 5, 15 percent partitions for training, validation and testing respectively.

We evaluate three finetuned BERT-like models: BERT, RoBERTa, XLNet along with GPT-3, GPT-3.5, and GPT-4. We evaluate each GPT model's zero-shot performance with prompting alone. For GPT-3.5, we also consider a finetuned variant trained with the same data as the BERT-like models. 

For sequence tagging in word-level detection, we apply the BIO tagging scheme to all words. During training, we also mark subword units with inside tokens when slang words are split into tokens. During evaluation, however, we only consider whole words splitted by white space to ensure a consistent evaluation metric across models with different tokenization schemes. 

For GPT models, we evaluate zero-shot performance by prompting the task instruction along with the input sentence. For sentence-level detection, we use the prompt:

\begin{enumerate}[label={>>>}, align=left]
    \item Is there a slang in the following sentence? Answer only 'Yes' or 'No'.
    \item Sentence: [A SENTENCE IN THE DATA]
    \item Answer:
\end{enumerate}
Similarly for word-level detection, we use the prompt:

\begin{enumerate}[label={>>>}, align=left]
    \item Identify slang in the following sentence. If a slang has been found, output the slang only. If no slang has been found, answer '[No slang]'.
    \item Sentence: [A SENTENCE IN THE DATA]
    \item Answer:
\end{enumerate}
For GPT-3.5 and GPT-4 under OpenAI's chat framework, we use the default system message ``You are a helpful assistant.'' for all prompts.

We mimic sequence labeling by searching for the resulting text span in the input sentence. When a match can be found, we set the appropriate beginning and inside labels for the detected span. We set the maximum number of generated tokens to be 1 and 20 for sentence-level and word-level detection respectively and set temperature to 0 for deterministic generation.

For finetuning GPT-3.5, we use '0' and '1' as binary labels for sentence-level detection. For word-level detection, we use the slang term as specified by the annotators.

\subsection{Slang source identification} \label{appex3}

We perform region identification on sentences from OpenSub-Slang with at least 2/3 annotator agreement, keeping those in which exact one copy of the slang can be found. We sample the sentences to construct an even sample of US and UK sentences which results in 1242 sentences for evaluation. We apply the same sampling scheme to sentences from the Green's Dictionary of slang for region and age identification. We obtain a sample of 6,096 sentences evenly partitioned across US and UK, as well as 4,008 sentences uniformly partitioned across six decades. We split all data samples into 80, 5, 15 percent partitions for training, validation and testing respectively. To determine whether a word is a content word, we refer to the set of English stop words in NLTK~\cite{bird04}.

We evaluate GPT-3's zero-shot performance on slang source identification by promoting the sentence:

\begin{enumerate}[label={>>>}, align=left]
    \item The following text is most likely produced in which region? Answer only 'US' or 'UK'.
    \item Text: [A SENTENCE IN THE DATA]
    \item Region:
\end{enumerate}
Similarly, we use the following prompt for age identification:

\begin{enumerate}[label={>>>}, align=left]
    \item Classify The following text into one of the following decades based on the language use. Possible answers include '1950', '1960', '1970', '1980', '1990', or '2000'. Answer in one word.
    \item Text: [A SENTENCE IN THE DATA]
    \item Decade:
\end{enumerate}
Similar to the slang detection prompts, we use the default system message ``You are a helpful assistant.'' for all GPT-3.5 and GPT-4 prompts.

We set the maximum number of generated tokens to be 1 for all slang source identification tasks and set temperature to 0 for deterministic generation.

We finetune GPT-3.5 using the same set of training and validation data used to train the BERT-like models. We use the same labels as in the zero-shot prompts as the target labels. This includes \{US, UK\} for regional identification and \{1950, 1960, 1970, 1980, 1990, 2000+\} for age identification.

\subsection{Probing model confidence} \label{appex2}

We use slang-containing sentences from OpenSub-Slang with a confidence score of 2 or above.
We use entries where both the slang word and its paraphrase tokenize into single tokens by all models\footnote{We observe that all models have the tendency to assign much higher probabilities to subword tokens, regardless of whether they are part of slang or literal tokens. See Appendix~\ref{apptoken} for a detailed analysis}. Since all GPT models are autoregressive language models, we truncate all tokens after the slang word for fair comparison and remove all sentences in which the slang appears at the beginning. After pre-processing, we obtain 220 sentence pairs from OpenSub-Slang. This includes 59 sentence pairs from US movies and 161 from UK movies.

\subsection{Preprocessing Green's Dictionary of Slang}

For each definition entry in Green's Dictionary of Slang (GDoS), we automatically extract usable usage context sentences from the entry's corresponding list of citations. For each citation, we apply a simple heuristic to extract potential example sentences by matching all text followed by a series of numeric characters and a column (e.g. ``212:''). For all extracted sentences, we ensure that the slang word can be found in the sentence after tokenizing by whitespace. This results in 33,181 entries with example sentences attached with a total of 99,181 example sentences. From the citation in which an example sentence was extracted from, we associate the corresponding date and region tags of the citation with the example sentence.

\begin{figure}[t!]
		\begin{subfigure}[b]{0.95\linewidth}
			\caption{Time - All}
			\includegraphics[width=\linewidth]{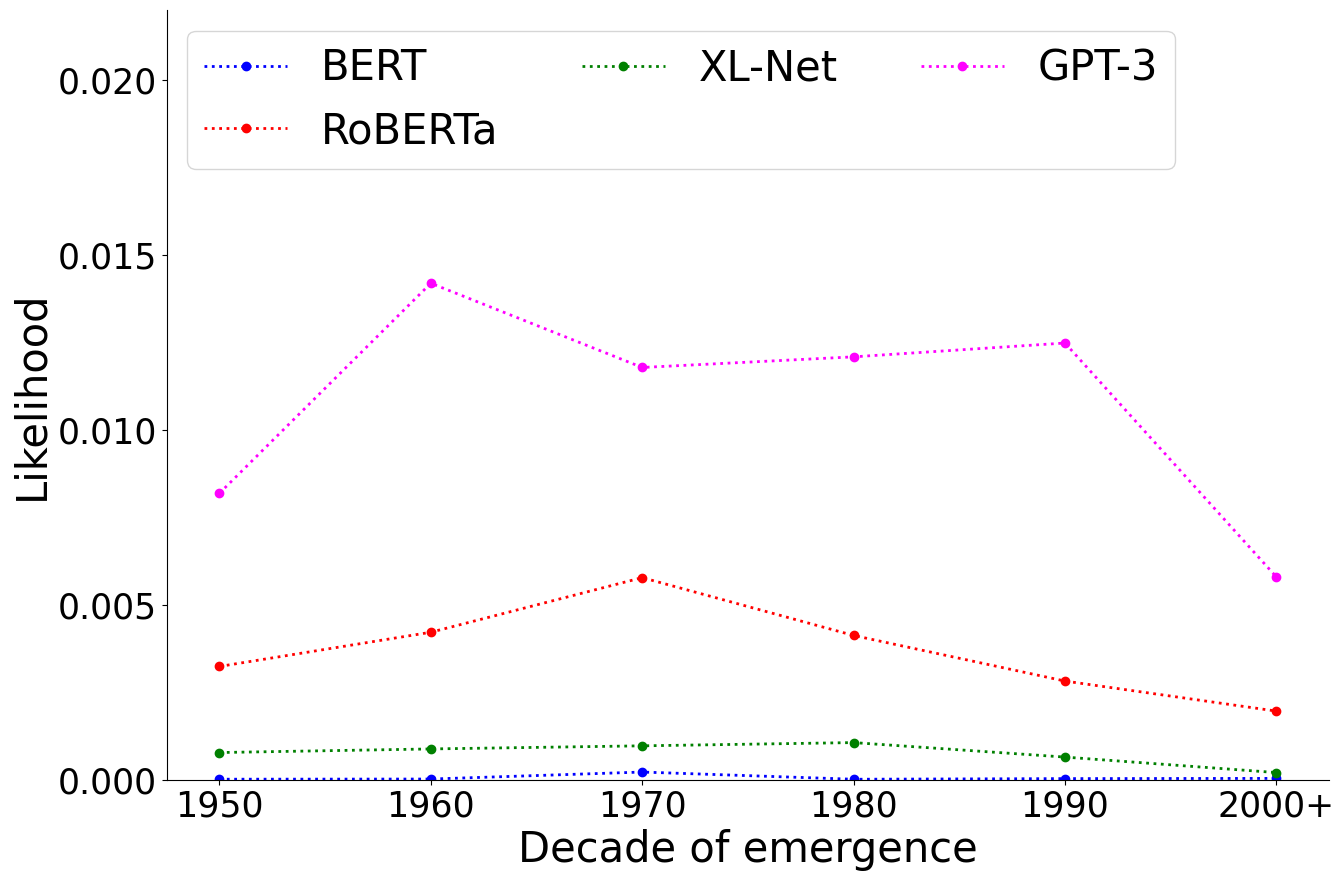}
		\end{subfigure}\
		\begin{subfigure}[b]{0.95\linewidth}
			\vspace{0.2cm}
			\caption{Time - US}
			\includegraphics[width=\linewidth]{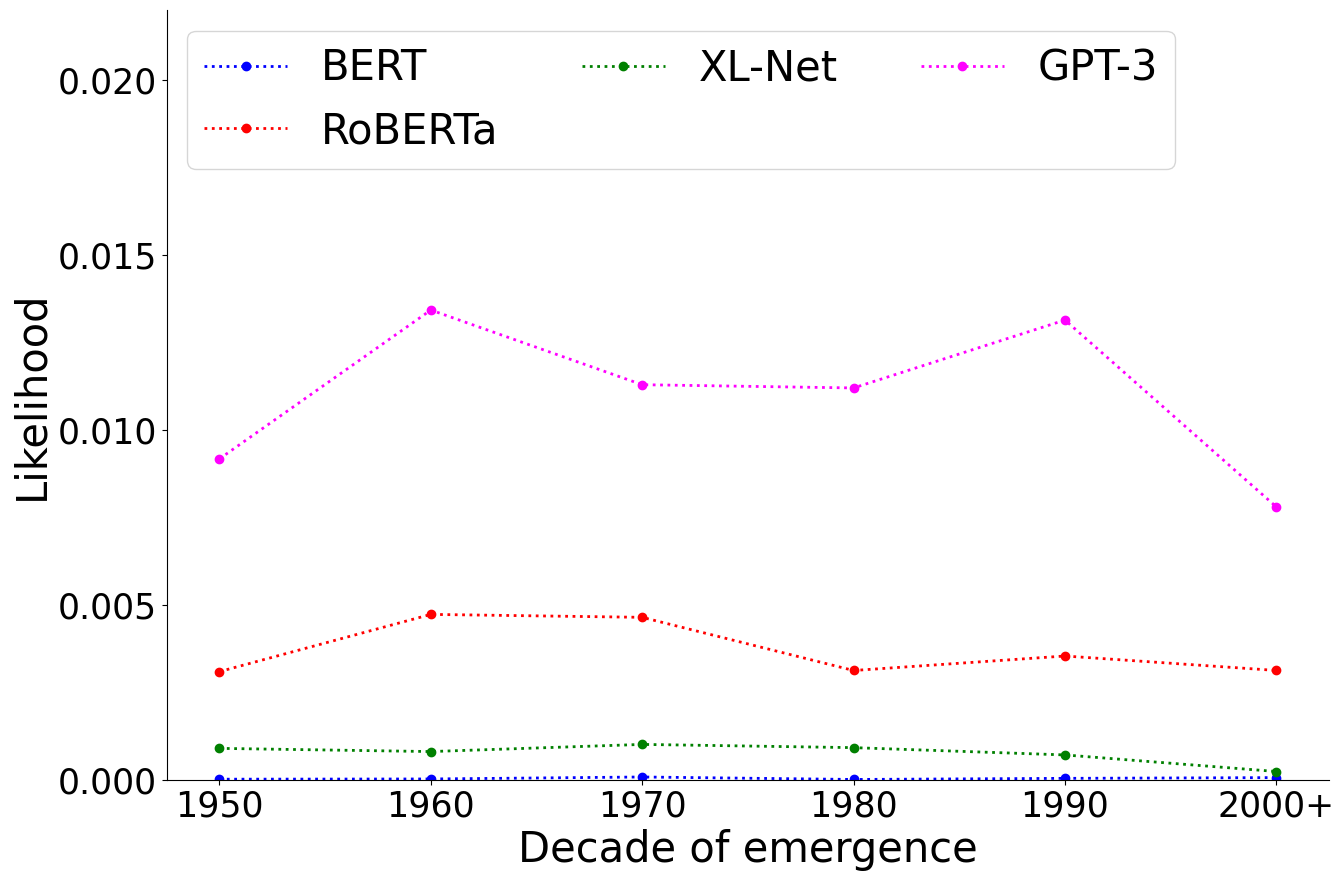}
		\end{subfigure}
        \begin{subfigure}[b]{0.95\linewidth}
			\vspace{0.2cm}
			\caption{Time - UK}
			\includegraphics[width=\linewidth]{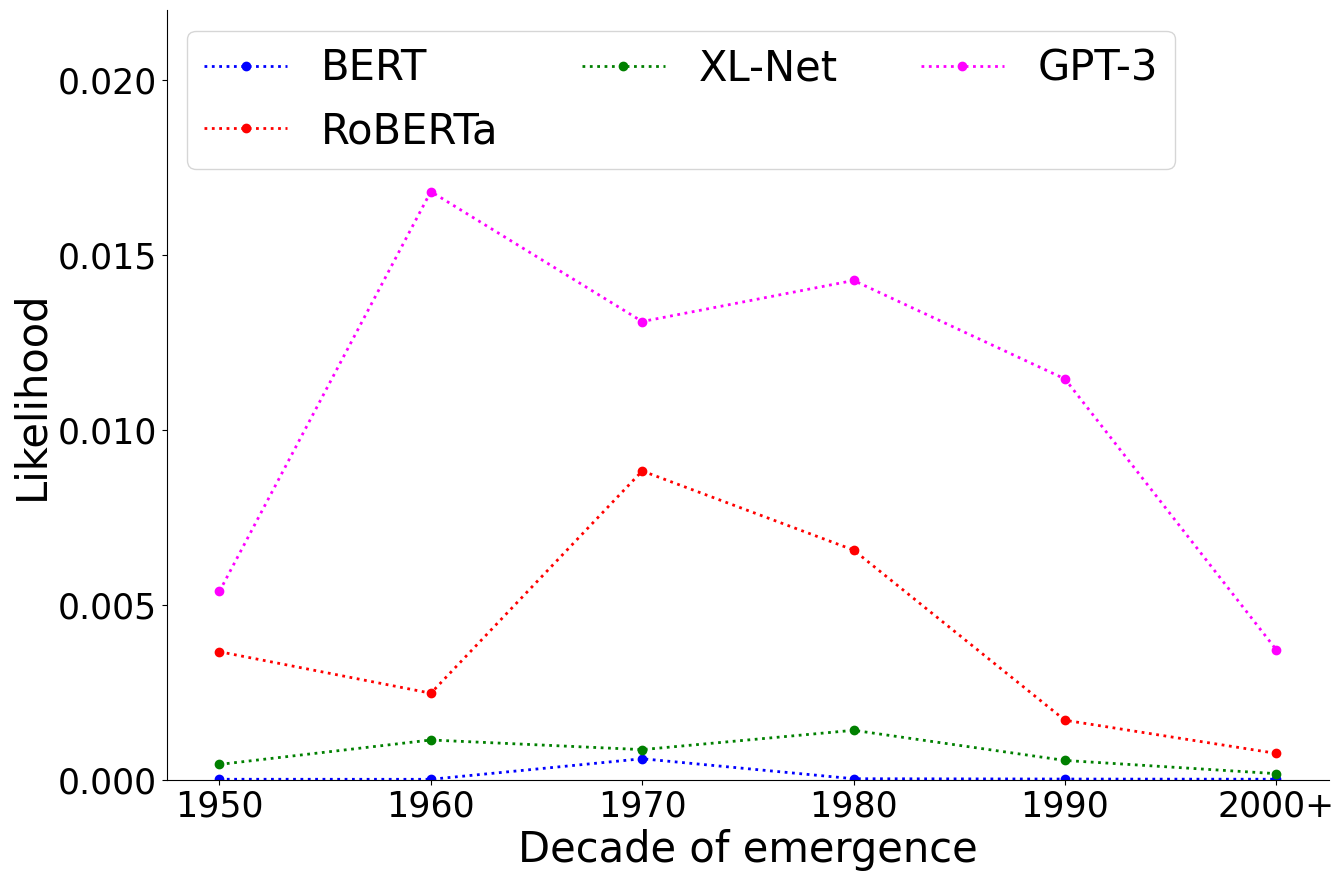}
		\end{subfigure}
		\caption{Mean LM probabilities over slang tokens in sentences across different time periods.} 
		\label{figmlmagegds}
	\end{figure}

 \begin{figure}[t!]
		\begin{subfigure}[b]{0.95\linewidth}
			\caption{Region - All}
			\includegraphics[width=\linewidth]{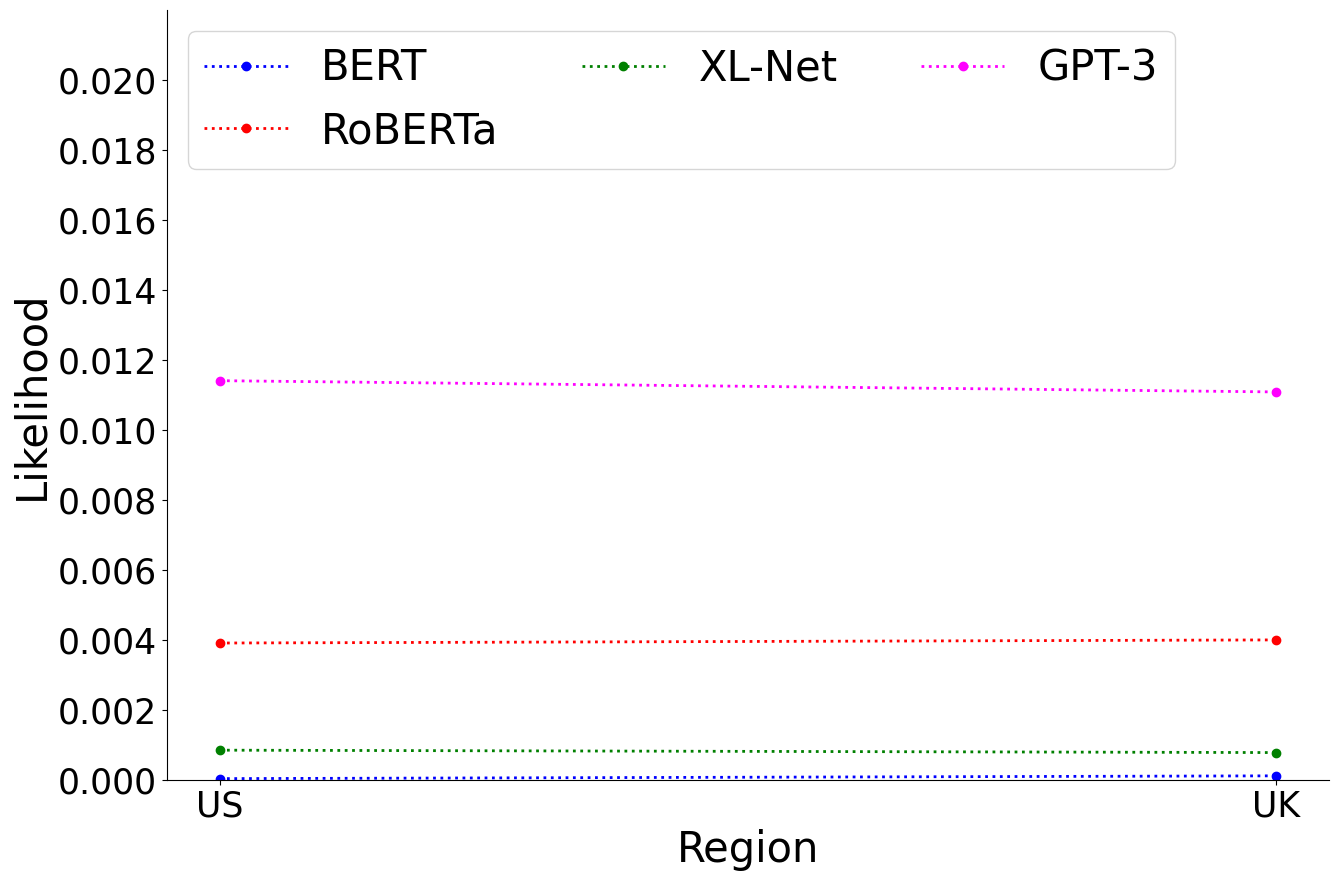}
		\end{subfigure}\
		\begin{subfigure}[b]{0.95\linewidth}
			\vspace{0.2cm}
			\caption{Region - Pre 1980}
			\includegraphics[width=\linewidth]{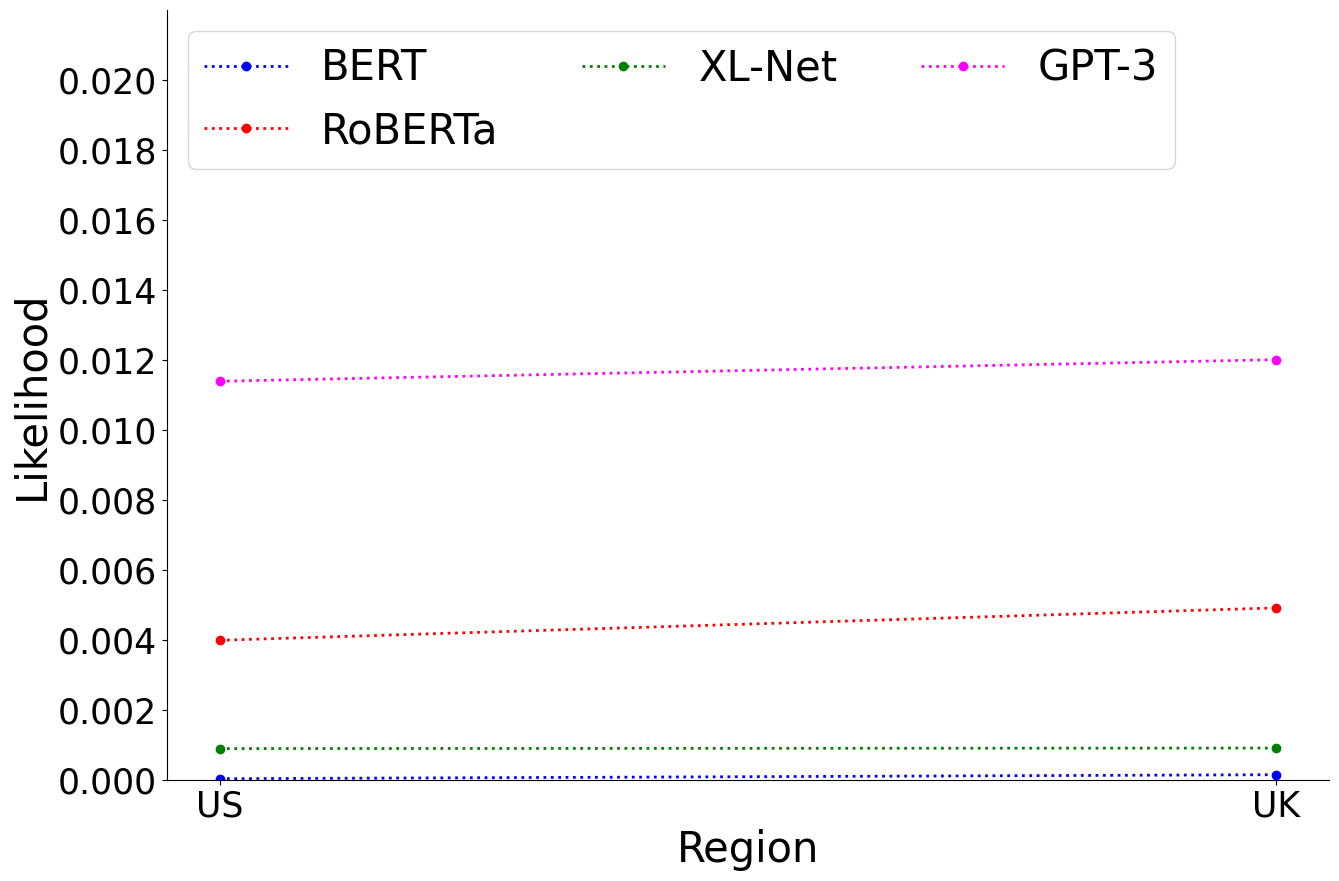}
		\end{subfigure}
        \begin{subfigure}[b]{0.95\linewidth}
			\vspace{0.2cm}
			\caption{Region - Post 1980}
			\includegraphics[width=\linewidth]{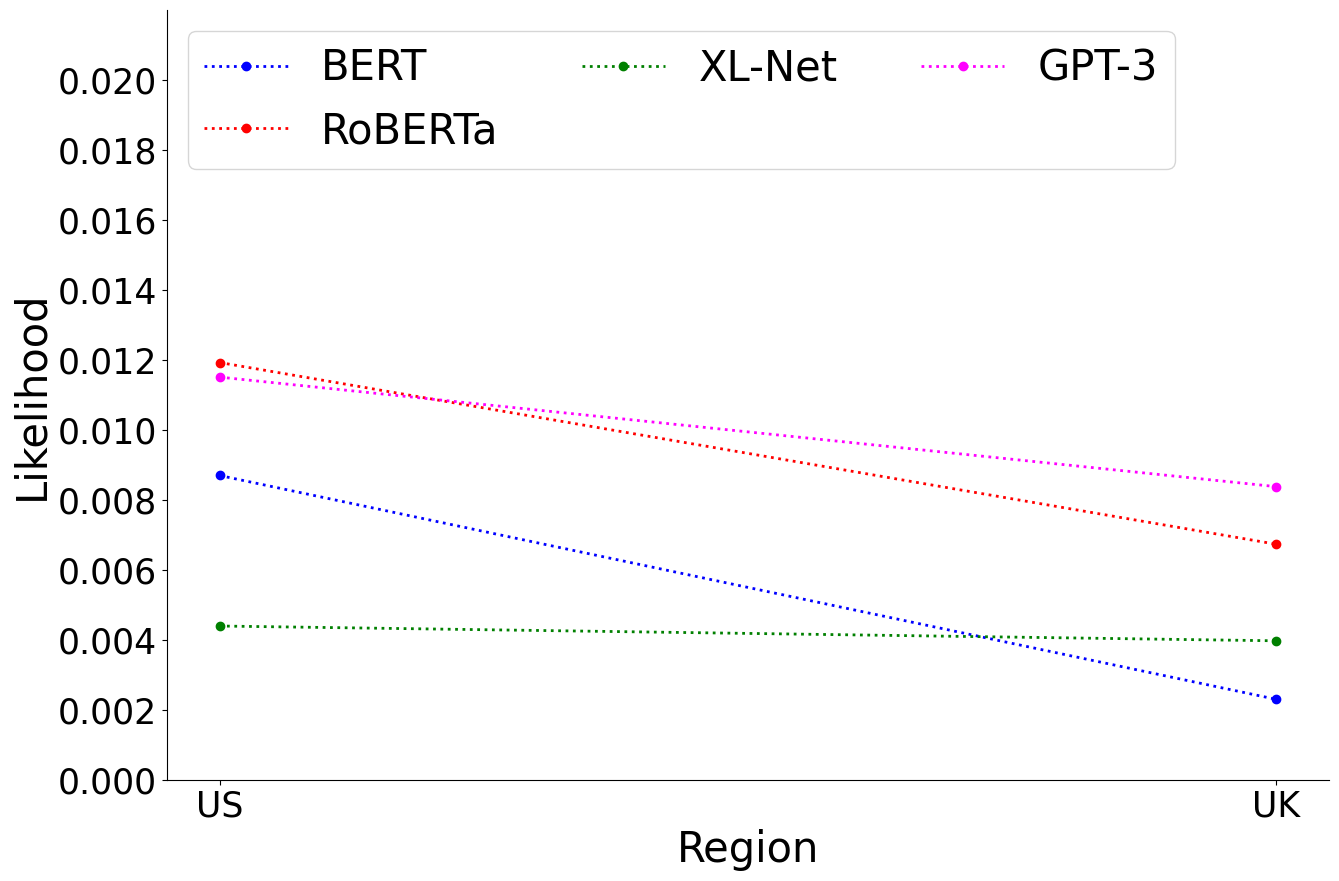}
		\end{subfigure}
		\caption{Mean LM probabilities over slang tokens in sentences across different regions.} 
		\label{figmlmreggds}
	\end{figure}

\section{Slang Token Probabilities over Time and Region} \label{appgreens}

We measure the language model probabilities assigned to slang tokens for entries in Green's Dictionary of Slang (GDoS). We use a similar task setup as described in Section~\ref{ex2}. However, since GDoS does not contain any literal paraphrases for the slang tokens, we only measure the absolute probabilities assigned to slang tokens. Here, we focus on how the models perform differently over different sets of slang usages stratified across both time and region.

We consider all example sentences in which the corresponding slang word can be represented using a single token by all models. Furthermore, we only consider sentences with a region tag of US or UK and a date tag after the year of 1950. This results in 5,052 sentences in total, with 3,617 sentences from the US and 1,435 from the UK. Of the 5,052 sentences, we have 1,285, 1,431, 859, 615, 564 sentences from each decade respectively from 1950s to 1990s and 298 sentences from the year 2000 and onwards.

Figure~\ref{figmlmagegds} shows the result over different time periods and Figure~\ref{figmlmreggds} over different regions. Overall, we observe consistent performance across different time periods and regions from all models. One exception to this is that for both BERT, RoBERTa, and GPT-3 the probabilities drop significantly for contemporary slang usages recorded after 1980. This is especially noticeable for slang usages from the UK. We postulate that the models likely make very little distinction for older slang from both regions, but for newer ones the models are exposed to slang usages from the US much more frequently than ones from the UK. We also observe that GPT-3 is a lot less confident on new slang usages from after the year 2000. These findings are consistent with our main results on the OpenSub-Slang dataset where all slang usages are extracted from movies produced after the year 2000.

\begin{table}[t!]
		\begin{subfigure}[b]{\linewidth}
			\centering
			\caption{BERT}
			\begin{tabular}{lrrr}
				\multicolumn{1}{l}{Tokenization} & Literal & Slang \\
				\hline
				\addlinespace[0.1cm]
				All Single &  0.00017 & 4.24e-05 \\
				\addlinespace[0.1cm]
				Slang: Single &  0.208 & 1.81e-05 \\
                Literal: Multiple \\
                \addlinespace[0.1cm]
				Slang: Multiple 0.000118 & 0.252 \\
                Literal: Single \\
			\end{tabular}
		\end{subfigure}\hfill
		
		\vspace{0.2cm}
		
		\begin{subfigure}[b]{\linewidth}
			\centering
			\caption{RoBERTa}
			\begin{tabular}{lrrr}
				\multicolumn{1}{l}{Tokenization} & Literal & Slang \\
				\hline
				\addlinespace[0.1cm]
				All Single &  0.0146 & 0.00555 \\
				\addlinespace[0.1cm]
				Slang: Single &  0.265 & 0.00454 \\
                Literal: Multiple \\
                \addlinespace[0.1cm]
				Slang: Multiple & 0.0111 & 0.29 \\
                Literal: Single \\
			\end{tabular}
		\end{subfigure}\hfill
		
		\vspace{0.2cm}

        \begin{subfigure}[b]{\linewidth}
			\centering
			\caption{XLNet}
			\begin{tabular}{lrrr}
				\multicolumn{1}{l}{Tokenization} & Literal & Slang \\
				\hline
				\addlinespace[0.1cm]
				All Single &  0.00163 & 0.000642 \\
				\addlinespace[0.1cm]
				Slang: Single &  0.118 & 0.00141 \\
                Literal: Multiple \\
                \addlinespace[0.1cm]
				Slang: Multiple & 0.00484 & 0.0831 \\
                Literal: Single \\
			\end{tabular}
		\end{subfigure}\hfill
		
		\vspace{0.2cm}
		
		\begin{subfigure}[b]{\linewidth}
			\centering
			\caption{GPT-3}
			\begin{tabular}{lrrr}
				\multicolumn{1}{l}{Tokenization} & Literal & Slang \\
				\hline
				\addlinespace[0.1cm]
				All Single &  0.0293 & 0.014 \\
				\addlinespace[0.1cm]
				Slang: Single &  0.285 & 0.0125 \\
                Literal: Multiple \\
                \addlinespace[0.1cm]
				Slang: Multiple & 0.0208 & 0.182 \\
                Literal: Single \\
			\end{tabular}
		\end{subfigure}\hfill
		\caption{Mean language model likelihood scores of slang and literal tokens under different tokenization conditions. The first row in each table shows the probability scores on sentences where both the slang and literal tokens are tokenized into single tokens by all models. The next two rows show results on sentences where the individual model tokenizes one word type with multiple tokens but uses a single token to represent the other.}
		\label{toktable}
	\end{table}

\section{Effect of Tokenization in Probing} \label{apptoken}

In our experiments shown in Section~\ref{ex2} and Appendix \ref{appgreens} involving probing probabilities, we only consider the subset of sentences in which both the corresponding slang and literal paraphrase (if applicable) can be presented using a single token. We perform this sampling procedure because we observe that all models tend to assign much higher probabilities to subword tokens. That is, regardless of whether a word is used as a slang or a literal paraphrase, words that comprise of subword tokens always attain much higher probabilities. Table~\ref{toktable} shows probabilities on slang containing sentences from OpenSub-Slang partitioned by tokenization. This is problematic as the tokenization scheme is dictating the magnitute of the probabilities over distributional semantics. We thus control for this confound by only considering sentences where all words of interest can be tokenized into a single token.
\end{document}